\documentclass[runningheads]{article}
\usepackage[T1]{fontenc}
\usepackage{graphicx}
\usepackage{caption} 
\usepackage{multirow}
\usepackage[table,xcdraw]{xcolor}
\usepackage{adjustbox}
\usepackage{authblk}
\usepackage{orcidlink}

\begin{document}
\title{Explainable Emotion Decoding\\ for Human and Computer Vision}

%

\author[2,4]{Alessio Borriero~\orcidlink{0009-0009-8850-090X}}
\author[1]{Martina Milazzo~\orcidlink{0009-0002-5246-8556}}
\author[2]{Matteo Diano~\orcidlink{0000-0001-7024-5183}}
\author[2,3]{Davide Orsenigo~\orcidlink{0009-0001-3085-562X}}
\author[2]{Maria Chiara Villa~\orcidlink{0000-0002-1617-6081}}
\author[2]{Chiara DiFazio~\orcidlink{0000-0002-8684-6714}}
\author[2,5]{Marco Tamietto~\orcidlink{0000-0002-8815-8499}}
\author[3]{Alan Perotti~\orcidlink{0000-0002-1690-6865}}

\affil[1]{La Sapienza University of Rome, Italy}
\affil[2]{University of Turin, Italy}
\affil[3]{Centai Institute, Turin, Italy}
\affil[4]{University of Camerino, Italy}
\affil[5]{Tilburg University, Netherlands}

%
%
%
\date{}
\maketitle              
\begin{abstract}

Modern Machine Learning (ML) has significantly advanced various research fields, but the opaque nature of ML models hinders their adoption in several domains. Explainable AI (XAI) addresses this challenge by providing additional information to help users understand the internal decision-making process of ML models. In the field of neuroscience, enriching a ML model for brain decoding with attribution-based XAI techniques means being able to highlight which brain areas correlate with the task at hand, thus offering valuable insights to domain experts.

In this paper, we analyze human and Computer Vision (CV) systems in parallel, training and explaining two ML models based respectively on functional Magnetic Resonance Imaging (fMRI) and movie frames. We do so by leveraging the "StudyForrest" dataset, which includes functional Magnetic Resonance Imaging (fMRI) scans of subjects watching the "Forrest Gump" movie, emotion annotations, and eye-tracking data.

For human vision the ML task is to link fMRI data with emotional annotations, and the explanations highlight the brain regions strongly correlated with the label. On the other hand, for computer vision, the input data is movie frames, and the explanations are pixel-level heatmaps.

We cross-analyzed our results, linking human attention (obtained through eye-tracking) with XAI saliency on CV models and brain region activations. We show how a parallel analysis of human and computer vision can provide useful information for both the neuroscience community (allocation theory) and the ML community (biological plausibility of convolutional models).\\

\end{abstract}
%
%
%


\section{Introduction}
Machine Learning (ML) algorithms have revolutionized data analysis by leveraging big data to extract valuable insights that might surpass and enhance human understanding. In the realm of emotion decoding, ML can be applied to process diverse data sources to unravel the complexities of human emotion, decoding the emotional content from various sensory sources like facial expressions, voice tones, and contextual cues ~\cite{ahmed_systematic_2023}. By integrating high-level semantic information, ML goes beyond traditional methods, enabling the analysis of emotional cues within natural scenes with sufficient accuracy ~\cite{lin_advancing_2023}.
Indeed, the construction of emotional perception in the human brain is intrinsically complex and mediated by multiple hierarchical and parallel layers of processing, ranging from visual low-level features to the complex reconstruction of affective semantics. Each affective information is processed through feedback loops that extensively involve the entire brain, from subcortical to cortical areas and vice versa; the identification of the brain regions linked to emotion representation remains an unresolved issue ~\cite{mitchell2012conscious}\cite{van2011cortico}. The current debate features a clash between two opposing theories regarding the neural correlates involved in the creation and perception of emotional feelings: the \textit{locationist} theory posits that discrete emotion categories consistently and specifically correspond to distinct brain regions, whereas the \textit{constructionist} approach argues that discrete emotion categories are constructed of more general brain networks not specific to those categories ~\cite{lindquist_brain_2012}.
In recent years, the application of ML techniques in neurophysiological studies regarding the decoding of emotional states from neuroimaging data (such as EEG and fMRI) has uncovered intriguing insights into the brain's processing of emotions~\cite{kragel_multivariate_2015}\cite{kragel_decoding_2016}. The classification of emotional states has always been a challenge for computer vision as well, because of the complex nature of emotional information, which does not rely uniquely on low level spatial features, but mainly on high level semantic information. Numerous studies have focused on applying artificial intelligence to the task of decoding facial expressions~\cite{Mellouk_Facial_2020}\cite{ko_brief_2018}, yet employing ML models to derive emotional information from natural scenes proves to be more challenging~\cite{Pikoulis_Leveraging_2021}\cite{Thuseethan_EmoSeC_2022}.
One of the central goals of this work is to examine side by side the processes by which the human brain and deep learning models decode emotional states from natural scenes, with the objective of properly comparing biological and artificial vision systems. The ambition to analyze the similiarities between human and artificial vision is motivated by many works, which have highlighted fundamental analogies between state-of-the-art computer vision algorithms and human brain visual system ~\cite{kriegeskorte_deep_2015}\cite{barrett_analyzing_2019}\cite{saxe_if_2021}. Specifically, it has been shown that the layers within Convolutional Neural Networks (CNNs) exhibit a similar structure to the brain's areas in the ventral stream visual pathways in terms of their internal representations~\cite{yamins_using_2016}\cite{schrimpf_brain-score_2018}\cite{lindsay_convolutional_2021}. However, little is known about the similarities and the differences regarding emotional decoding of human and machines. Does artificial intelligence generate similar inner representations of emotional visual input with respect to human beings?   
In order to pursue this research direction, we leveraged XAI techniques~\cite{arrieta_survey}. Since modern ML models (including CNNs) offer no human-understandable representation of their inner decision logic, XAI tackles this problem by providing additional information with respect to a ML model decision; in this work we rely on two canonical attribution methods, LIME~\cite{ribeiro_why_2016} and SHAP~\cite{lundberg_unified_2017}. When dealing with brain decoding models, the explanation scores highlight the brain areas strongly correlated with the task at hand, while for CV models the explanation heatmaps provide saliencies at pixel level on movie frames.
To explore the complexity of the human emotional spectrum, we exploited a dataset derived from fMRI acquisitions of subjects viewing the "Forrest Gump" movie, namely the StudyForrest project~\cite{hanke_high-resolution_2014}\cite{sengupta_studyforrest_2016}. Although the dataset came with emotional annotations~\cite{labs_portrayed_2015}, we chose to use the annotations provided by Lettieri et al. in their work about emotionotopy in the human brain~\cite{lettieri_emotionotopy_2019}.

The present work aims to address the problem of the neural correlates related to emotion processing in the human brain exploiting machine learning models and XAI techniques. Moreover, merging eyetracking data, XAI results from brain decoding models and the explanation obtained from the emotion decoding on movie frames, we tried to bridge CV models with the human visual system, looking for neural patterns correlating with the degree of attention match between CNNs and human beings.

\section{Related works}
\subsection{Explainable Computer Vision}
Since Alexnet~\cite{alexnet}, Deep Learning models have been the de-facto architecture for Computer Vision. However, these models offer no human-understandable representation of their inner decision processes, behaving like black boxes. The research field of XAI tackles this problem, developing approaches that help human understand the behaviour of black-box models. Since there is no mathematical definition of explainability and interpretability, it is important to keep in mind that this concept is linked only to the human understanding of neural network decision process: in~\cite{Miller_Explanation_2019} interpretability is explained as "the degree to which a human can understand the cause of a decision".

XAI is therefore a broadly defined concept which is implemented through a plethora of competing algorithms~\cite{bodria_benchmarking_2021}. 
There are fundamental differences that help categorise these approaches: (i) {\em ante-hoc} vs {\em post-hoc}, (ii) {\em local} vs {\em global}, and (iii) {\em model-agnostic} vs {\em model-aware}.
Ante-hoc approaches try to directly train explainable models, while post-hoc approaches aim at explaining an already existing opaque model. Local explanations pertain to a single data-point (e.g. {\em why was this image classified as deepfake?}), while global explanations try to extract a general description of the black-box model as a whole (e.g. {\em how does this deepfake detector work?}). Finally, model-aware explainers leverage the internals of a black-box, for instance computing scores based on gradients, while model-agnostic explainers only require to query a black-box at will. Most explainers are post-hoc and local, and there exist both model-aware and model-agnostic explainers for CV models, but the two most used ones, LIME~\cite{ribeiro_why_2016} and SHAP~\cite{lundberg_unified_2017}, are model-agnostic.

LIME~\cite{ribeiro_why_2016} (Local Interpretable Model-Agnostic Explanations) focuses on perturbing the single input sample, querying the model on the newly obtained synthetic point cloud in order to observe the impact on the output, and training a white-box surrogate on the labelled point cloud. For computer vision tasks, LIME segments the input image into superpixels (patches of pixels) and exploits masking as a perturbation function.



SHAP (Shapley Additive exPlanations)~\cite{lundberg_unified_2017}
exploits the concept of Shapley value coming from cooperative game theory. The Shapley approach was to consider the power-set of players in a team, in order to measure the impact of each single player in the team outcome. SHAP keeps the same mindset, substituting players with features and team outcome with black-box output. Like LIME, there is a fundamental activity of perturbation/masking, but SHAP computes importance scores through marginalisation, without training a surrogate model.


\subsection{Brain decoding: Machine Learning on fMRI data}

Brain decoding refers to the process of interpreting both exogenous and endogenous brain states from observable brain activities, taking brain activity as input and brain states as output~\cite{k_deep_2015}. 
Concerning fMRI experiments, in which brain activity is measured through an indirect estimation of the metabolic changes in blood flow~\cite{heeger_what_2002}, decoding analysis has been traditionally performed through single-voxel univariate methods, like general linear model (GLM)~\cite{connolly_representation_2012}. This type of approach is referred to as ‘univariate’ because the corresponding statistical tests only consider the value of a single voxel or ROI (Region Of Interest) per condition at a time. Recently, an increasing number of researchers are adopting analyses that focus on patterns of responses across multiple voxels, known as multivariate pattern analysis (MVPA), instead of relying on values from single voxels or regions, in order to better assess the highly nonlinear information processing in the human brain~\cite{haxby_multivariate_2012}\cite{weaverdyck_tools_2020}. MVPA techniques can be implemented using straightforward correlation analysis, linear classifiers, partial least squares algorithms~\cite{lee_fast_2022} or by employing traditional machine learning algorithms, such as Support Vector Machines~\cite{kubilius_brain-decoding_2015}, with the most prominent examples of MVPA leveraging deep neural networks~\cite{firat_deep_2014}\cite{k_deep_2015}.

Compared to the CV case, less research work from the XAI field has focused on brain decoding tasks:
except for a few valuable examples mostly regarding diagnostic brain imaging~\cite{farahani_explainable_2022}~\cite{pat_explainable_2023}, fMRI and brain decoding studies lack of application of explainability approaches. With the present work we aimed to explore the possibility to exploit XAI to explore and study complex brain states. 

\subsection{Emotion decoding for human and Computer Vision}
Numerous studies in CV have tackled the problem of emotion decoding, with a primary focus on decoding emotions from facial expressions~\cite{haines_using_2019}. While some studies categorize emotion through a limited set of basic emotional states (happiness, sadness, surprise, anger, fear, and disgust)~\cite{fasel_automatic_2003}, others decompose emotions into fundamental dimensions like valence and arousal~\cite{gunes_emotion_2011}. State-of-the-art emotion decoding models exploit CNNs~\cite{lopes_facial_2017}, while recent works analyze dynamic emotional facial expressions in videos~\cite{du_spatio-temporal_2021}.

Since fMRI data provide extensive insights into high-level cognitive processes in the human brain, numerous studies have focused on the task of decoding emotions through machine learning algorithms utilizing this type of neuroimaging data~\cite{kragel_decoding_2016}\cite{kragel_multivariate_2015}. The state-of-the-art literature has demonstrated the possibility to predict at least information about emotional dimensions (valence and arousal) of emotional feelings perceived by human beings~\cite{baucom_decoding_2012}\cite{heinzle_multivariate_2012}.

\section{Experimental setup}

\subsection{Frames, fMRI and labels}

\subsubsection{The StudyForrest project}
The keystone of our study is the StudyForrest project \cite{hanke_high-resolution_2014} \cite{hanke_simultaneous_2016}. This work is a scientific initiative that involves the collection and analysis of neuroimaging data, particularly functional magnetic resonance imaging (fMRI) data, related to dynamic visual and auditory stimuli, in order to study various aspects of human cognition and brain functions. "Forrest Gump" is a movie directed by Robert Zemeckis and released in 1994. Recognized as a classic of american cinema, the complexity of its narrative lay the foundation for our exploration into the emotional dynamics portrayed throughout its duration. It was shot at a standard cinematic frame rate of 23.97 frames per second (fps) The resolution, adhering to the standards of cinema, is 16:9. In total, the film comprises 204501 frames. For the ML+XAI pipelines described below, we adopted the movie version described of the original StudyForrest work: 8 movie segments, for a total of 172405 frames and an overall duration of 120 minutes, in order to divide the fMRI acquisition of each subject in 8 runs.

\subsubsection{Emotion annotation}
\begin{figure}[h!]
  \includegraphics[width=\linewidth]{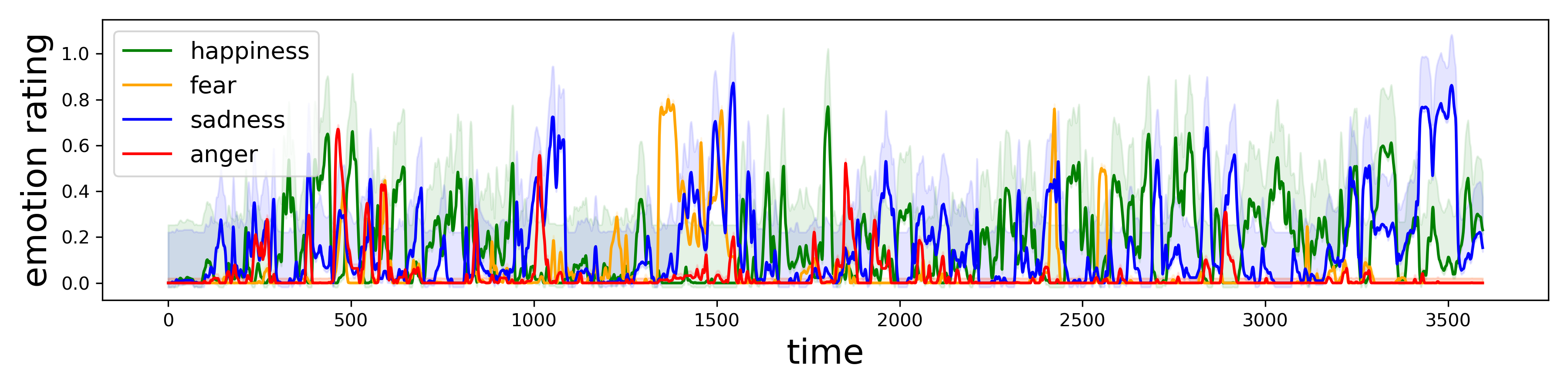}
  \caption{Time series of emotion annotations. The work by Lettieri et al. provides the emotion annotation by 12 indipendent human annotators of the whole Forrest Gump movie; in our work we focused on happiness, fear, sadness and anger.}
  \label{fig:emotion_labeling}
\end{figure}

An important component of this work is the inclusion of an emotional labeling, which we use as ground-truth labels for both the CV-ML model and the fMRI-based brain decoder. The StudyForrest project provides emotional labeling, but we decided to exploit the emotion annotations by Lettieri et al. \cite{lettieri_emotionotopy_2019} that has enriched
the original dataset with the inclusion of labels related to the emotional responses associated with the fMRI. In this work 12 healthy Italian native speakers participants have been instructed to rate the movie with respect of 6 basic emotions (happiness, surprise, fear, sadness, anger and disgust) during the same reduced version of Forrest Gump used for the StudyForrest project. Annotators were allowed to report more than one emotion at the same time and ratings were continuously  recorded. Therefore, six different times series for each of the twelve subjects has been produced. A temporal resampling has been implemented, in order to align the emotion data with the fMRI data in terms of temporal resolution. In this paper we focused on the four fundamental emotions of happiness, fear, sadness, and anger, as depicted in Figure~\ref{fig:emotion_labeling}.
In order to binarize our labels dataset and determine whether an emotion was dominant or not, a criterion was established. An emotion was considered dominant if at least one annotator assigned it a value greater than the other emotions at a
given moment. If an emotion was dominant for one annotator at a specific instant, it was considered positive; otherwise, it was considered negative.
This binarization process allowed us to transform the continuous emotion intensity annotations into a binary representation, providing a clear distinction between dominant and non-dominant emotions. Regarding the brain decoding task, a further preprocessing step has been implemented before the binarization procedure. A sliding-windows smoothing has been applied over the emotion annotations time-series, with a windows size equal to 10 second and a stride equal to 2 seconds.

Through the procedure described above we built 5 different binary dataset, 4 for the emotions and 1 for the face, which were provided to 5 different ML models. The choice to have different models for different emotions lies on the theoretical prior of our work. One of our aims has been to investigate the dualism between constructionist and locationist theories of emotion processing in the human brain. By definition of the model itself, a single multi-class model would not allow us to see shared networks among different emotions, resulting in an imbalance towards the locationist vision.

\subsubsection{Movie frames datasets}
The first preprocessing step aimed to build the frames datasets for the emotion decoding task has been to resample the whole movie according to the fMRI repetition time (TR), namely 2 seconds. Thus, we obtained a set of 3599 frames.
However, we observed that some subsequent frames were identical or nearly-identical, especially for long, static scenes. We wanted to remove duplicates in order to avoid a train-test leakage. We first tried to do so by measuring pixel-level difference between consecutive frames, but this approach did not capture semantical similarities: for instance, in scenes with crops blowing in the wind, we had high-pixel level differences - within a semantically monotone scene. In order to select significative frames, 
we resorted to exploit an Imagenet-pretrained CNN, specifically the EfficientNet B0 model \cite{tan_efficientnet_2019}. 
For all frames, the EfficientNet was employed to derive labels indicative of the elements present in the input image. From these labels, the top 3 values among the 1000 provided by the network were selected. Subsequently, for each pair of consecutive frames, the indices of the obtained labels were compared, leading to the elimination of frames that shared one or two labels. This process
effectively removed frames that exhibited similarities in their content, enhancing the diversity and relevance of the dataset. 

Alongside the emotion decoding task, in this work we aimed to asses the problem of facial recognition. To do that, we exploited DeepFace \cite{serengil_lightface_2020}, an open-source library which embeds face recognition and facial expressions decoding. Through a specific function, we were able to extract bounding boxes around faces in individual frames. The binary labels were then selected based on the following criteria: frames without faces were labelled as negative cases, while frames with 1 or 2 faces with a bounding box's area greater than or equal to 4 \% of the whole frame were labelled as positive cases.
In order to adapt our movie frames to the requirements of the CNN, which demands square images, we applied ad hoc cropping and padding to the whole collection of frames.
Since our frames datasets were highly imbalanced, we performing an undersampling procedure.
To conclude, after the preprocessing steps described above, we obtained 5 binary datasets of movie frames: 4 related to the 4 basic emotions and the last concerning the presence of faces within frames.

\subsubsection{The fMRI dataset}
For the brain decoding task we exploited the rich fMRI dataset provided by StudyForrest, where 15 German native subjects participated to fMRI recording sessions, during which they were engaged in watching Forrest Gump. All subjects underwent 8 fMRI runs, for a total o 120 minutes of acquisition. The data preprocessing pipeline was prepared using AFNI~\cite{cox_software_1997} and FSL (FMRIB Software Library)~\cite{jenkinson_fsl_2012} software. Structural images were resampled to 1mm3 (\textit{3dAllineate}). Then they were brain extracted (\textit{standardspaceroi; bet2}), corrected for intensity bias (\textit{3dUnifize}), and spatially normalized to the Montreal Neurological Institute (MNI) space with non-linear registration (\textit{3dQwarp}). All functional volumes were slice timing corrected (\textit{3dTshift}), spatially realigned to the first volume of the functional acquisition and corrected for scan motion (\textit{3dvolreg}). All functional volumes were then spatially smoothed (\textit{3dBlurToFWHM}) with a 6mm full-width half-maximum isotropic Gaussian kernel (FWHM) and the signal was normalized (center: 0; variations in percentage). Each average EPIs were aligned to correspondent high resolution T1w and then re-sampled in the size of the functional acquisition using a weighted sinc-interpolation method. After the standard preprocessing pipeline described above, we introduced a lag of 2 seconds to account for the delay in hemodynamic activity. The resulting timeseries were then temporally smoothed using a moving average procedure (10 seconds window). This method allowed us to further account for the uncertainty of the temporal relationship between the actual onset of emotions and the time required to report the emotional state. As for the movie's frames dataset, we applied undersampling to our dataset in order to assess the imbalance in the class distribution, for each of the 5 datasets.

\subsubsection{Eyetracking}
In addition to the fMRI acquisition, the StudyForrest project also provides an eyetracking registration for each subject during each session. Such data enriches our analysis with an insightful information about the dynamics of the human attention of each subject with respect to the visual stimulus. In our study, we exploited the normalized version of the eye movement recordings, sampled uniformly at 1000 Hz, with a spatial resolution of 1280x546 pixels, the same as the original movie.

\subsection{Machine Learning on movie frames}
In order to decode emotional states from movie frames, we exploited a transfer learning approach~\cite{zhuang_comprehensive_2021}. We first chose a pretrained CNN, namely the EfficientNet B0~\cite{tan_efficientnet_2019}, then we added a customized layer on the top of the pre-trained network, allowing the model to learn features related to the binary emotion decoding and face/no-face tasks. The pre-training
endowed the model with the ability to capture general features and patterns present in images, while the subsequent training phase allowed it to adapt to the variations of our domain-specific task. The additional layer was made by a global average pooling 2D layer, a dropout layer and dense layer (1024 neurons) and an output neuron implementing a sigmoid function for the classification. A key step of the fine-tuning procedure has been a grid-search over the hyper-parameters of the tailored layer. We explored the following space of hyperparameters: number of units in dense layers (128, 256, 512), dropout rate (0.1, 0.3, 0.5), learning rate (0.01, 0.001, 0.0001) and batch size (8, 16, 32). Moreover, we included in the grid-search two possible choices as the first layer, namely a global average pooling 2D layer or a flatten layer. To avoid overfitting and improve the model's generalizability, we relied on early stopping~\cite{yao_early_2007} in the fine-tuning procedure. Finally, to enrich the original datasets with the aim to further enhance the generalization capabilities of our models, we performed data augmentation~\cite{shorten_survey_2019}, applying random flip, random rotation and random zoom to the frames. For what concerns the train-test splitting, we relied on a standard leave-one-out k-fold procedure, splitting each of the datasets in 5 folds. The results shown in the experimental results section are the average accuracy over the 5 splits.

\subsection{Machine Learning on fMRI data}
In the context of the brain decoding task, the goal is infer information about dominant emotions and face occurrences from fMRI data. A key feature of our work has been to parcel each brain volume according to a brain atlas, namely the one by Glasser et al.~\cite{glasser_multi-modal_2016}. Thus, each input of the ML models consists in a set of 394 features, representing the mean activity of each area of the brain in a given moment of the fMRI acquisition.
We chose a within-subject approach, training and testing one model for each of the experimental subjects provided by the StudyForrest database. Our choice of training subject-specific models is due to the high variability, noisy nature and high dimensionality of the BOLD activity from different brains, which makes the generalization of the fmri signal across-subject one of the main issues in brain decoding tasks ~\cite{akamatsu_perceived_2021}. Many solutions have been explored to address the possibility to obtain subject-independent models, such as hyperalignment ~\cite{liang_cross-subject_2020} and roi-specific models ~\cite{yousefnezhad_supervised_2021}. In our work, we tried to apply a simple leave-one-subject-out approach, without the exploitation of more sophisticated techniques, achieving bad results in terms of generalization performance. Considering our goal of performing a whole-brain analysis, we then decided to implement a subject-wise paradigm, which was able to provide a sufficient amount of  information about the structure of the processing of high-level semantic information within the human brain.
We built a multi-layer perceptron~\cite{glorot_understanding_nodate} model, with relu as activation function, a single output neuron employing a sigmoid function and Adam algorithm as optimizer~\cite{kingma_adam_2017}. A grid-search procedure has been implemented in order to choose the best set of hyperparameters among the following: number of hidden layer (1 or 2) and the numbers of neurons for each layer (ranging from 40 to 300). In order to enforce generalization capabilities of the models, the learning algorithm has been equipped with a regularization parameters. As for the computer vision models, we performed a leave-one-out k-fold splitting, with k=5, averaging the resulting accuracy values over the 5 splits. 

\subsection{XAI for emotion decoding}
\begin{figure}[h!]
  \includegraphics[width=\linewidth]{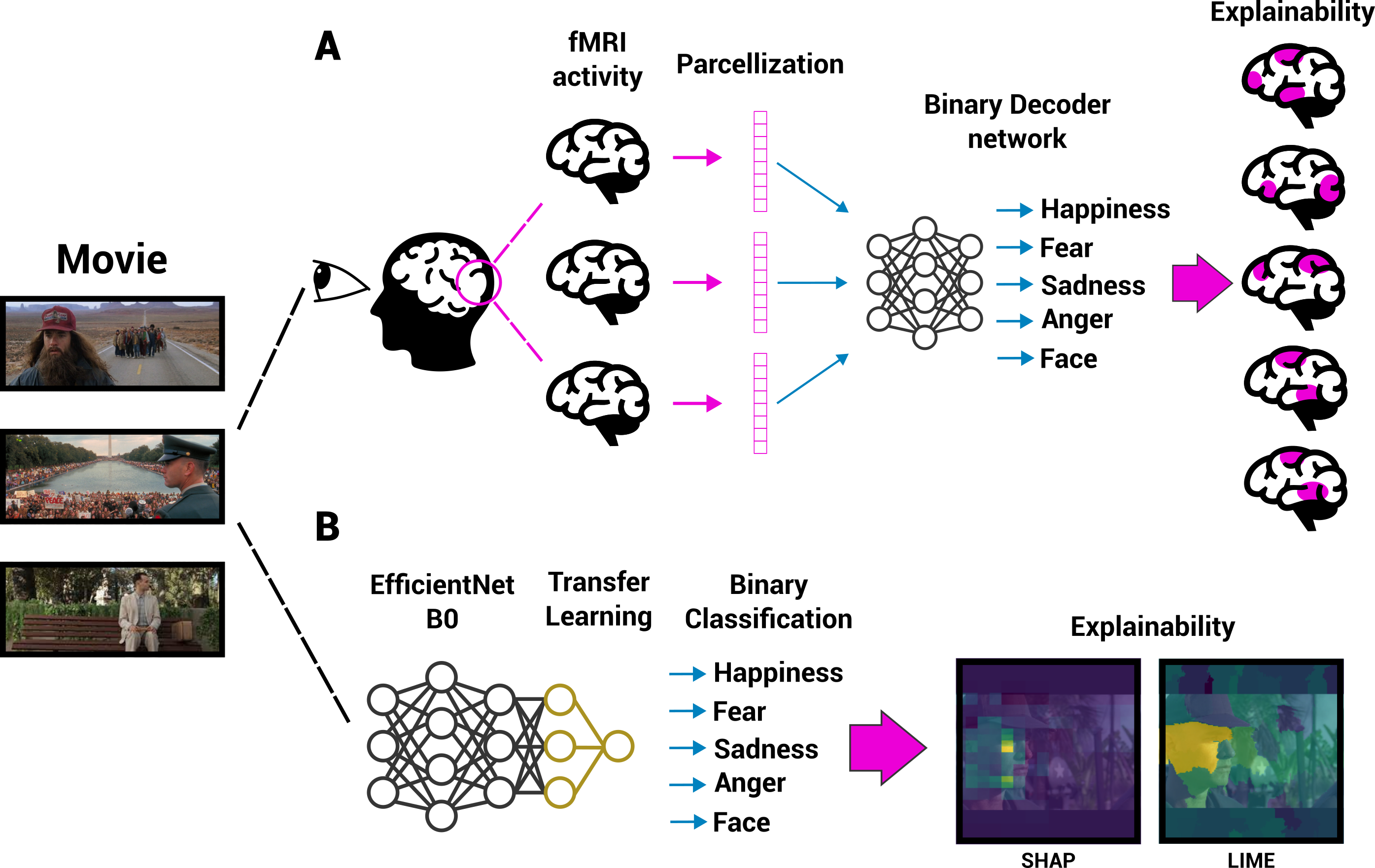}
  \caption{Emotion decoding and XAI pipelines for brain data (A) and computer vision (B).}
  \label{fig:brain_decoding_pipeline}
\end{figure}

For both our set of emotion and face decoding models, namely the CV and the fMRI-based models, we exploited LIME and SHAP as XAI techniques, in order to unveil the decision-making processes behind the 'black-box'. An overview of these two ML+XAI pipelines is visualized in Figure~\ref{fig:brain_decoding_pipeline}.
Since exploring different explainability methods in brain decoding problems wasn’t the aim of our work we do not exploit other XAI methods, such as gradient-based ones. Thus, we prefer to adopt the most used model-agnostic approaches.

For CV models, the XAI methods we applied generated heatmaps across frames, assigning significance scores to different areas of each frame. Specifically, for each frames of each dataset (regarding the four binary emotion plus face classification) we generated an explanatory heatmap, giving us an importance score for each pixel regarding the class discrimination task.

For what concerns the brain decoding models, we exploited XAI techniques in order to assign a score of importance to each area of the brain, namely the features for our ML problem. This approach produces a brain map of area importances for each sample of our dataset, i.e. the brain volumes of the fMRI acquisition.

In order to asses the statistical validity of the XAI brain maps that we obtained applying SHAP and LIME to our brain decoding models, we compute a null model. To do so, we trained the ML model with a shuffled set of binary labels, to assess chance-level performance. The set of feature importance values obtained by explaining the null model are used to build a null distribution, which in turn we will use to validate our explanation scores.

\subsection{CNN-humans attentional match: a comparative analysis}
A crucial step of our work has been the cross-analysis between explainable CV and fMRI XAI results, alongside with the information provided by the eyetracking. Due to the temporal synchronization of all the different domains of our analysis, we could compare the visual attention of the CV model and of the human beings involved in the experiment. Our basic aim has been to identify which brain areas process the emotional content of the movie when the CNN and human look at the same spot within the visual stimulus. To do so, we first define a frame-wise score of overlap of visual attention. Exploiting the XAI heatmaps, we set this overlap score as the percentile of the heatmap's value for which the eyetracking recording fall within the pixels with importance values above that percentile. In order to provide more robustness to our analysis and to asses the timing of the human visual attention process, we consider a window of 1 second around the given frame when measuring the overlap score. Once we computed this measure for each of the movie's frames, we performed an area-wise correlation between overlap score of each frame and importance value of the given area at the time of each frame. This analysis provides a whole-brain map that describes which brain areas contain more information about the emotional content of the visual stimulus whenever the human subject and the computer vision model focus their attention on the same spot within the movie frame.

\section{Experimental results}

\subsection{Machine Learning on movie frames}

Our investigation into emotion recognition through computer vision involved training distinct models for four key emotions (happiness, sadness, fear, and anger)
and one for face detection. In order to optimize the performance of our emotion recognition models, we employed the grid search method, as mentioned in the experimental setup section. As reported in Table \ref{tab:accuracies_cv}, each model shows good in-sample and out-of-sample performance. Specifically, the classifier designed to determine whether happiness is present in the frames exhibits higher accuracy compared to other emotion-related models.

For the face detection task, we obtained a remarkably higher performance with respect to the emotion classification models. This result is certainly due to the lower complexity of the face recognition problem, which relies on low-level visual features. On the contrary, the emotion decoding problem lives in a multi-faced semantic space making the identification of the visual features which define the presence of a certain emotion in a visual scene a non-trivial issue.

\begin{table}[h!]
\centering
\captionsetup{skip=10pt} 
\caption{Computer vision models performance}
\begin{tabular}{|l|l|l|l|l|l|}
\hline
\textbf{Model}                  & \textit{\textbf{Happiness}} & \textit{\textbf{Fear}} & \textit{\textbf{Sadness}} & \textit{\textbf{Anger}} & \textit{\textbf{Face}} \\ \hline
\rowcolor[HTML]{EFEFEF} 
\textbf{In-sample Accuracy}     & 0.99                        & 0.99                   & 0.98                      & 0.99                    & 0.99                   \\
\textbf{Out-of-Sample Accuracy} & 0.84                        & 0.80                   & 0.76                      & 0.76                    & 0.96                   \\ \hline
\end{tabular}
\label{tab:accuracies_cv}
\end{table}

\subsection{Machine Learning on fMRI data}

To assess the emotion and face decoding problem with fMRI data we trained 5 classifiers for each of the 15 experimental subjects. The overall out-of-sample performance we obtained, displayed in Table \ref{tab:accuracies_fmri}, is appreciably above the chance level. Moreover, the results seem stable across the subjects. Among the 5 models, the one related to face recognition shows the lowest performance. Such a result can be justified by a complex brain representation related to the face processing within the visual stimuli with respect to the emotional content of the movie. Moreover, the information about the faces rely only on visual stimuli, while the emotional nuances of the movie are carried by both visual and auditory stimuli. Thus, the emotion decoders can rely on a richer set of information throughout the brain with respect to the face decoder model. However, we analyzed in depth the brain representations related to each of the 5 decoder models through explainability techniques, as discussed in the following section.

\begin{table}[h!]
\centering
\captionsetup{skip=10pt} 
\caption{Brain decoding models performance}
\begin{adjustbox}{width=\textwidth}
\begin{tabular}{|l|l|l|l|l|l|}
\hline
\textbf{Model}                  & \textit{\textbf{Happiness}} & \textit{\textbf{Fear}} & \textit{\textbf{Sadness}} & \textit{\textbf{Anger}} & \textit{\textbf{Face}} \\ \hline
\rowcolor[HTML]{EFEFEF} 
\textbf{In-sample Accuracy}     & 0.976               & 0.971          & 0.945             & 0.970           & 0.925          \\
\textbf{Out-of-Sample Accuracy} & 0.911               & 0.923          & 0.887             & 0.894           & 0.797          \\ \hline
\end{tabular}
\end{adjustbox}
\label{tab:accuracies_fmri}
\end{table}

\subsection{Explainability for fMRI-based models}
\begin{figure}[h!]
  \includegraphics[width=\linewidth]{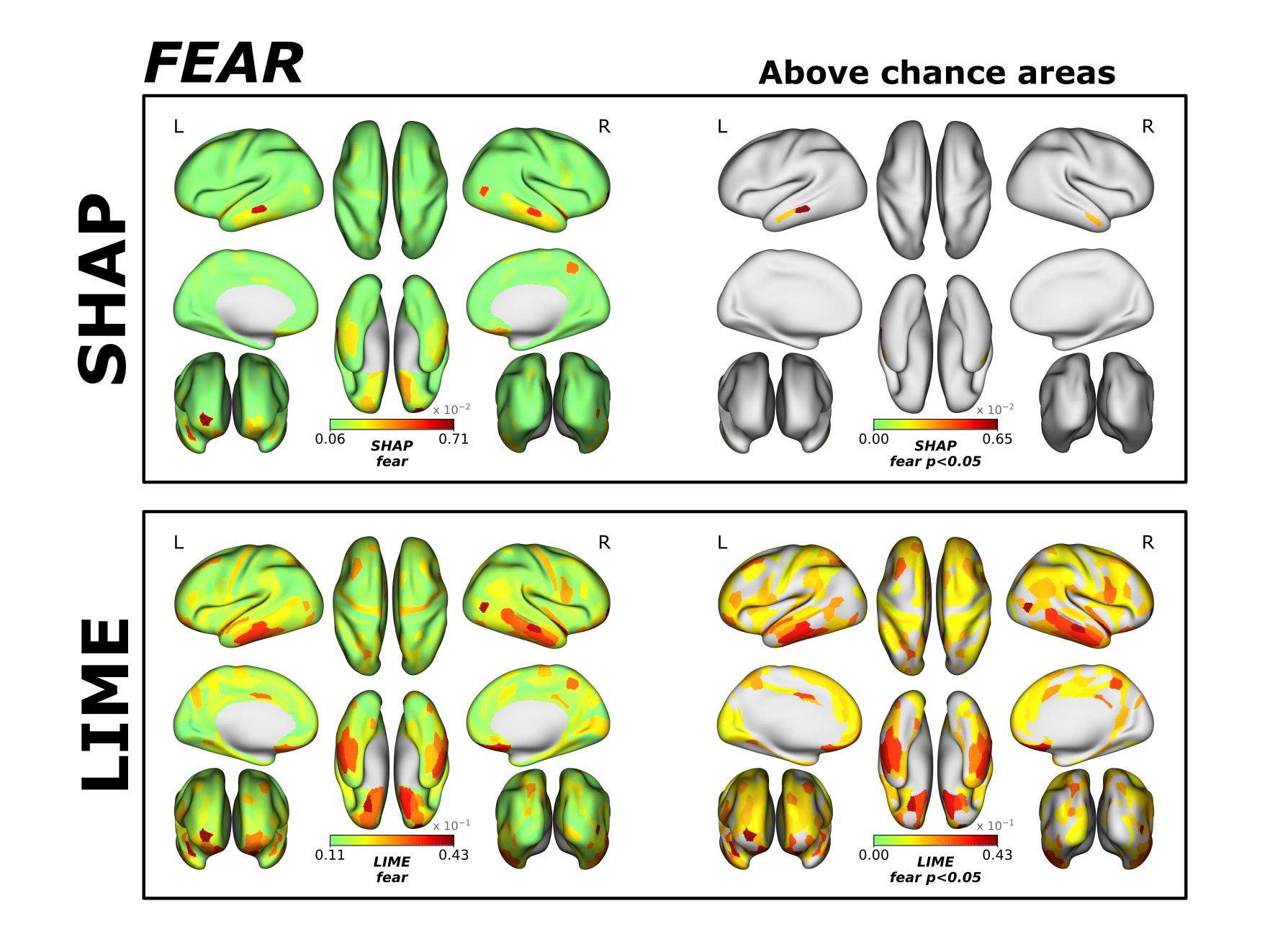}
  \caption{Brain-wise feature importance maps obtained with SHAP and LIME. Through a null model we assess the significance of each area, obtaining a limited set of regions which process most information about the emotional content of the movie.}
  \label{fig:ShapLime_brains}
\end{figure}

One of the most intriguing aspects of our study has been the evaluation of the features' importance within the brain decoding models. The application of SHAP and LIME as explainers for our fMRI-based classifiers lead us to unveil the most important brain areas in terms of emotion and face processing, by defining a set of brain networks related to each of the 5 models. Figure \ref{fig:ShapLime_brains} shows a representation of both SHAP and LIME brain maps related to the fear model.
Table \ref{tab:shap_limes_areas} summarizes the most significant areas for each of the four basic emotions and the face recognition, with respect to a null model.

\begin{table}[h!]
\centering
\captionsetup{skip=10pt} 
\caption{Most significant brain regions with respect to feature importance related to each brain decoding model.}
\begin{adjustbox}{width=\textwidth}
\begin{tabular}{|l|ll|ll|}
\hline
\rowcolor[HTML]{FFFFFF} 
\textbf{}                                                             & \multicolumn{2}{c|}{\cellcolor[HTML]{FFFFFF}\textbf{SHAP}} & \multicolumn{2}{c|}{\cellcolor[HTML]{FFFFFF}\textbf{LIME}} \\ \hline
\rowcolor[HTML]{FFFFFF} 
\textbf{Model}                                                        & \textbf{Macro Area}                                       & \textbf{Value}   & \textbf{Macro Area}                                          & \textbf{Value} \\
\rowcolor[HTML]{EFEFEF} 
\cellcolor[HTML]{EFEFEF}                                              & \cellcolor[HTML]{FCFF2F}Early Auditory Cortex     & 0.0027 & Insular Cortex                                     & 0.053 \\
\rowcolor[HTML]{EFEFEF} 
\cellcolor[HTML]{EFEFEF}                                              &                                                   &        & Ventral Stream Visual Cortex                       & 0.035 \\
\rowcolor[HTML]{EFEFEF} 
\cellcolor[HTML]{EFEFEF}                                              &                                                   &        & Orbital and Polar Frontal Cortex                   & 0.032 \\
\rowcolor[HTML]{EFEFEF} 
\cellcolor[HTML]{EFEFEF}                                              &                                                   &        & \cellcolor[HTML]{FCFF2F}Early Auditory Cortex      & 0.031 \\
\rowcolor[HTML]{EFEFEF} 
\multirow{-5}{*}{\cellcolor[HTML]{EFEFEF}\textit{\textbf{Happiness}}} &                                                   &        & Dorsal Stream Visual Cortex                        & 0.030 \\ \hline
\rowcolor[HTML]{FFFFFF} 
\cellcolor[HTML]{FFFFFF}                                              & \cellcolor[HTML]{FCFF2F}Lateral Temporal Cortex   & 0.0065 & Orbital and Polar Frontal Cortex                   & 0.043 \\
\rowcolor[HTML]{FFFFFF} 
\cellcolor[HTML]{FFFFFF}                                              &                                                   &        & Insular Cortex                                     & 0.041 \\
\rowcolor[HTML]{FFFFFF} 
\cellcolor[HTML]{FFFFFF}                                              &                                                   &        & \cellcolor[HTML]{FCFF2F}Lateral Temporal Cortex    & 0.040 \\
\rowcolor[HTML]{FFFFFF} 
\cellcolor[HTML]{FFFFFF}                                              &                                                   &        & Early Auditory Cortex                              & 0.028 \\
\rowcolor[HTML]{FFFFFF} 
\multirow{-5}{*}{\cellcolor[HTML]{FFFFFF}\textit{\textbf{Fear}}}      &                                                   &        & Superior Parietal Cortex                           & 0.027 \\ \hline
\rowcolor[HTML]{EFEFEF} 
\cellcolor[HTML]{EFEFEF}                                              & \cellcolor[HTML]{FCFF2F}Early Auditory Cortex     & 0.0044 & Orbital and Polar Frontal Cortex                   & 0.053 \\
\rowcolor[HTML]{EFEFEF} 
\cellcolor[HTML]{EFEFEF}                                              & \cellcolor[HTML]{FCFF2F}Lateral Temporal Cortex   & 0.0035 & Auditory Association Cortex                        & 0.039 \\
\rowcolor[HTML]{EFEFEF} 
\cellcolor[HTML]{EFEFEF}                                              &                                                   &        & \cellcolor[HTML]{FCFF2F}Lateral Temporal Cortex    & 0.037 \\
\rowcolor[HTML]{EFEFEF} 
\cellcolor[HTML]{EFEFEF}                                              &                                                   &        & \cellcolor[HTML]{FCFF2F}Early Auditory Cortex      & 0.030 \\
\rowcolor[HTML]{EFEFEF} 
\multirow{-5}{*}{\cellcolor[HTML]{EFEFEF}\textit{\textbf{Sadness}}}   &                                                   &        & Dorsolateral Prefrontal Cortex                     & 0.019 \\ \hline
\rowcolor[HTML]{FFFFFF} 
\cellcolor[HTML]{FFFFFF}                                              & \cellcolor[HTML]{FCFF2F}Superior Parietal Cortex  & 0.0039 & Insular Cortex                                     & 0.048 \\
\rowcolor[HTML]{FFFFFF} 
\cellcolor[HTML]{FFFFFF}                                              & Auditory Association Cortex                       & 0.0038 & Lateral Temporal Cortex                            & 0.030 \\
\rowcolor[HTML]{FFFFFF} 
\cellcolor[HTML]{FFFFFF}                                              & \cellcolor[HTML]{FCFF2F}Premotor Cortex           & 0.0038 & \cellcolor[HTML]{FCFF2F}Premotor Cortex            & 0.030 \\
\rowcolor[HTML]{FFFFFF} 
\cellcolor[HTML]{FFFFFF}                                              &                                                   &        & \cellcolor[HTML]{FCFF2F}Superior Parietal Cortex   & 0.027 \\
\rowcolor[HTML]{FFFFFF} 
\multirow{-5}{*}{\cellcolor[HTML]{FFFFFF}\textit{\textbf{Anger}}}     &                                                   &        & Orbital and Polar Frontal Cortex                   & 0.024 \\ \hline
\rowcolor[HTML]{EFEFEF} 
\cellcolor[HTML]{EFEFEF}                                              & Lateral Temporal Cortex                           & 0.0032 & Orbital and Polar Frontal Cortex                   & 0.034 \\
\rowcolor[HTML]{EFEFEF} 
\cellcolor[HTML]{EFEFEF}                                              & \cellcolor[HTML]{FCFF2F}Anterior Cingulate Cortex & 0.0022 & \cellcolor[HTML]{FCFF2F}Anterior Cingulate Cortex  & 0.019 \\
\rowcolor[HTML]{EFEFEF} 
\cellcolor[HTML]{EFEFEF}                                              &                                                   &        & Auditory Association Cortex                        & 0.017 \\
\rowcolor[HTML]{EFEFEF} 
\multirow{-4}{*}{\cellcolor[HTML]{EFEFEF}\textit{\textbf{Face}}}      &                                                   &        & Superior Parietal Cortex                           & 0.013 \\ \hline
\end{tabular}
\end{adjustbox}
\label{tab:shap_limes_areas}
\end{table}

At first, looking at the similarity among the two different explanatory maps obtained with SHAP and LIMES for each of the 5 models, we observed a good consistency in terms of resulting areas. As a stronger clue for this robustness with respect to XAI techniques, we compute pairwise spearman correlation corrected for a spin permutation test \cite{alexander-bloch_testing_2018} among the couples of explanation for each classification task, resulting in a strong correlation for all the models (Spearman R values: Happiness - 0.81, Fear - 0.83, Sadness - 0.85, Anger - 0.82,  Face - 0.81).

\begin{figure}[h!]
  \includegraphics[width=\linewidth]{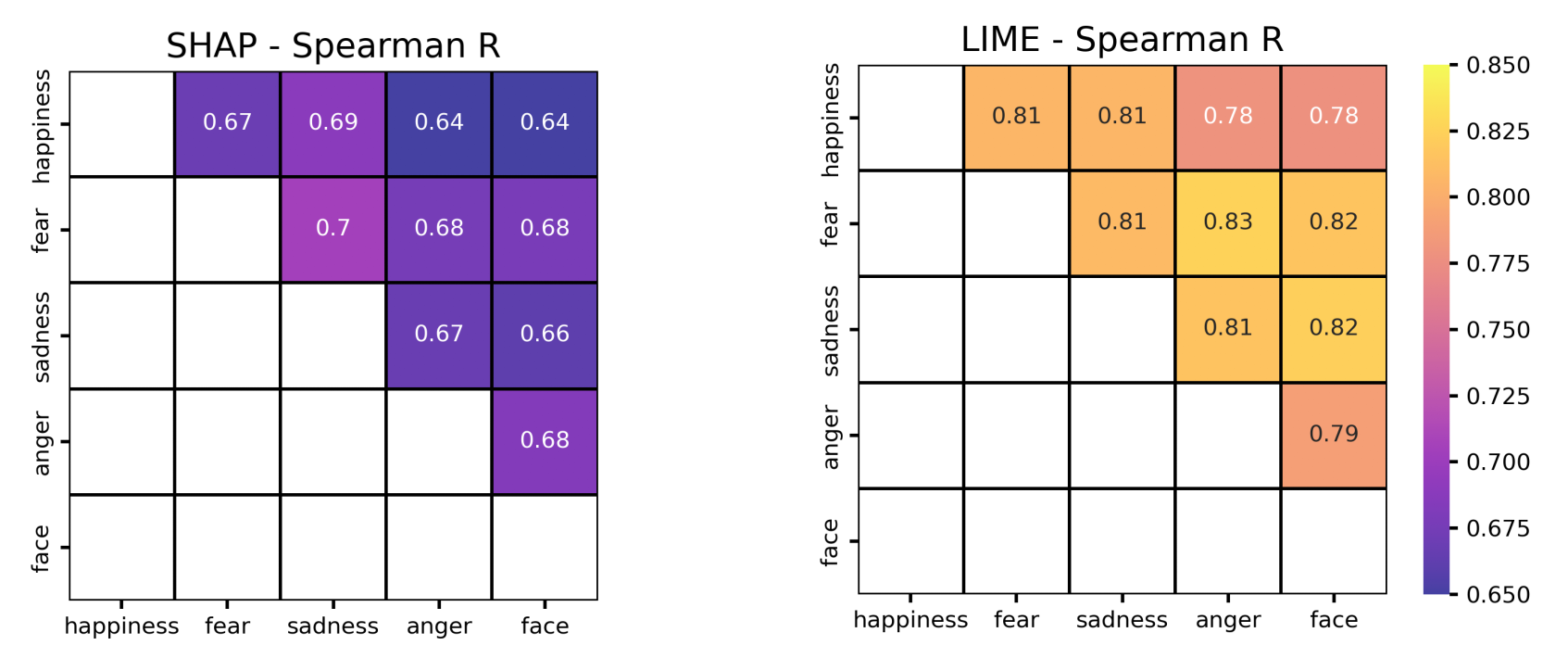}
  \caption{Correlation among brain maps related to different models. The high correlation values we observed are due to the existence of a common brain network which processes information about the emotional content of a multisensory input. All the resulting correlations have a strong statistical significance, with p-values always below than 0.0002.}
  \label{fig:correlation_among_emotion_bra}
\end{figure}

The feature importance maps we obtained provide interesting elements in context of the debate regarding how the brain processes different emotions. Looking at the overall pairwise Spearman correlation among the different maps corrected for a spin permutation test, (Figure \ref{fig:correlation_among_emotion_bra}), we observe a high correlation level among all the emotions, including the face-related map. This result underlines the existence of a common brain network for the processing of all the basic emotions we studied through our analysis. In particular, the most significant region represented in all the 5 explanation maps is the Orbital and Polar Frontal Cortex (OPFC). Historically, OPFC plays a crucial role with respect to the locationist hypothesis regarding the brain representation of the anger \cite{murphy_functional_2003} \cite{vytal_neuroimaging_2010}. Nevertheless, our results provide strong clues regarding a constructionist description of how emotional feelings and perceptions are represented in the human brain. From a psychological constructionist perspective, it is theorized that parts of the Orbital Frontal Cortex (OFC) contribute to core affect by serving as a hub for merging exteroceptive and interoceptive sensory data, thereby influencing behavior \cite{lindquist_brain_2012}. With the term core affect we describe the mental representation of bodily changes
that are sometimes experienced as feelings of hedonic pleasure
and displeasure with some degree of arousal \cite{barrett_chapter_2009}. 
Sensory information from the external environment and the body collectively direct an organism's reactions to its surroundings, allowing it to make actions that are appropriately adjusted to the context. Multiple connections involving the OFC to both various sensory systems and regions controlling visceral functions make this brain structure anatomically equipped to fulfill this function effectively \cite{kringelbach_functional_2004}. Thus, OFC serves as an high-level decision making core, which produces behavioral and cognitive response to emotional stimuli.

Our results depict more complex pictures than a single brain network involved in the processing of all the emotions. In fact, beyond the OFPC, other areas emerge in more than one model. Looking at the feature importance maps that we obtained, the Insula Cortex seems to convey much information about the emotional processing in the human brain, specifically for happiness, fear and anger. From a locationist perspective, the Insular cortex, in particular the Anterior Insula, is more involved in processing of emotional content, indicating a transition from sensory to emotional processing within this area, and representing the core of the sensation of disgust \cite{jabbi_common_2008}. The insula's intricate connections with other brain regions and its ability to integrate interoceptive inputs make it a key player in representing emotional experiences and bodily reactions to stimuli, as its activation has been linked to subjective feelings and cognitive factors, influencing decision-making processes and behavioral responses. Indeed, a more refined hypothesis, based on a constructionist approach, has identified a more complex role of this region in representing core affective feelings in awareness \cite{bud_craig_how_2009}. A
core feature in the mental states labeled “disgust” is
a representation of how an object will affect the body. In support of a psychological constructionist
view, Insula activation emerges in a number
of tasks that involve awareness of body states, shedding light on how specific aspects of insular activity can drive accurate emotional state predictions, but also on its potential inhibitory role in emotional processing.

Another notable brain area which emerges from our analysis is the Lateral Temporal Cortex, resulting as an important region for the classification models related to fear, sadness and anger. Looking with more details at our results, we find a peak of importance values in the Inferior Temporal Gyrus, namely the (bilateral) TE and TF areas. The state-of-the art literature describes these brain structures as a key step in the ventral visual stream implicated in object, face, and scene perception \cite{conway_organization_2018}. 

Alongside the Lateral Temporal Cortex's role in processing the visual aspects of emotions, it's important to highlight the significance of auditory characteristics in depicting emotional states. In this regard, it is crucial to mention the emergence of the Early Auditory Cortex as a significant area with respect to the feature importance score assigned by the explainers. Such a brain region can be observed in the models related to happiness, fear and anger. 

Thus, the results just discussed depict a well-defined emotion processing network which conveys most of the information about the emotional content of a complex visual and auditory stimulus. This set of brain regions seems to show a precise hierarchical structure. The OFPC represents a high-level cognitive stage, which integrate outer stimuli and inner representations influencing decision-making and modulating behavioral responses. At a mid-level, the Insular cortex acts as a relay for the aware response to sensory inputs, which can drive an emotional response. Finally, at the lower level, visual and auditory areas, namely the Lateral Temporal Cortex and the Early Auditory Cortex, represent a first integration step for the multisensory features conveying the emotional information.

Finally, looking at the neural correlates obtained from our analysis associated with face-related visual stimuli, we basically observe the interplay of two brain regions. The OPFC seems to play an important role in the processing of human faces. As discussed before, this area represents a high-level stage in the processing of emotional stimuli, merging inner and outer signals and providing a behavioral output. Alongside this cognitive step in the processing pathway we are trying to discuss, the Anterior Cingulate Cortex (ACC) seems to process a significant amount of information regarding the presence of a face in a natural visual scene, as it is involved in the processing of various types of face stimuli, including social and self-related faces \cite{morita2008role}. ACC represents a crucial hub in the human brain, integrating information at various levels through  connection with both the “emotional” limbic system and the “cognitive” prefrontal cortex \cite{stevens_anterior_2011}.
Interestingly, many works have highlighted the role of the ACC in processing face stimuli, both social or self-related \cite{hornak_changes_2003} \cite{morita_anterior_2014}. 
In particular, this region plays a significant role in the rapid processing of emotionally salient facial expressions, such as fear, thus providing an immediate emotional assessment of faces; additionally, the ACC's connections with various brain regions, including the amygdala and temporal cortical areas, facilitate its role for interpreting facial expressions and understanding social cues \cite{fan2011involvement} . The medial prefrontal network, which includes the ACC, is involved in processing emotional and social information from faces, indicating the ACC's role in social cognition and emotional regulation \cite{rolls2019cingulate}.

\subsection{Comparative analysis}

\begin{figure}[h!]
  \includegraphics[width=\linewidth]{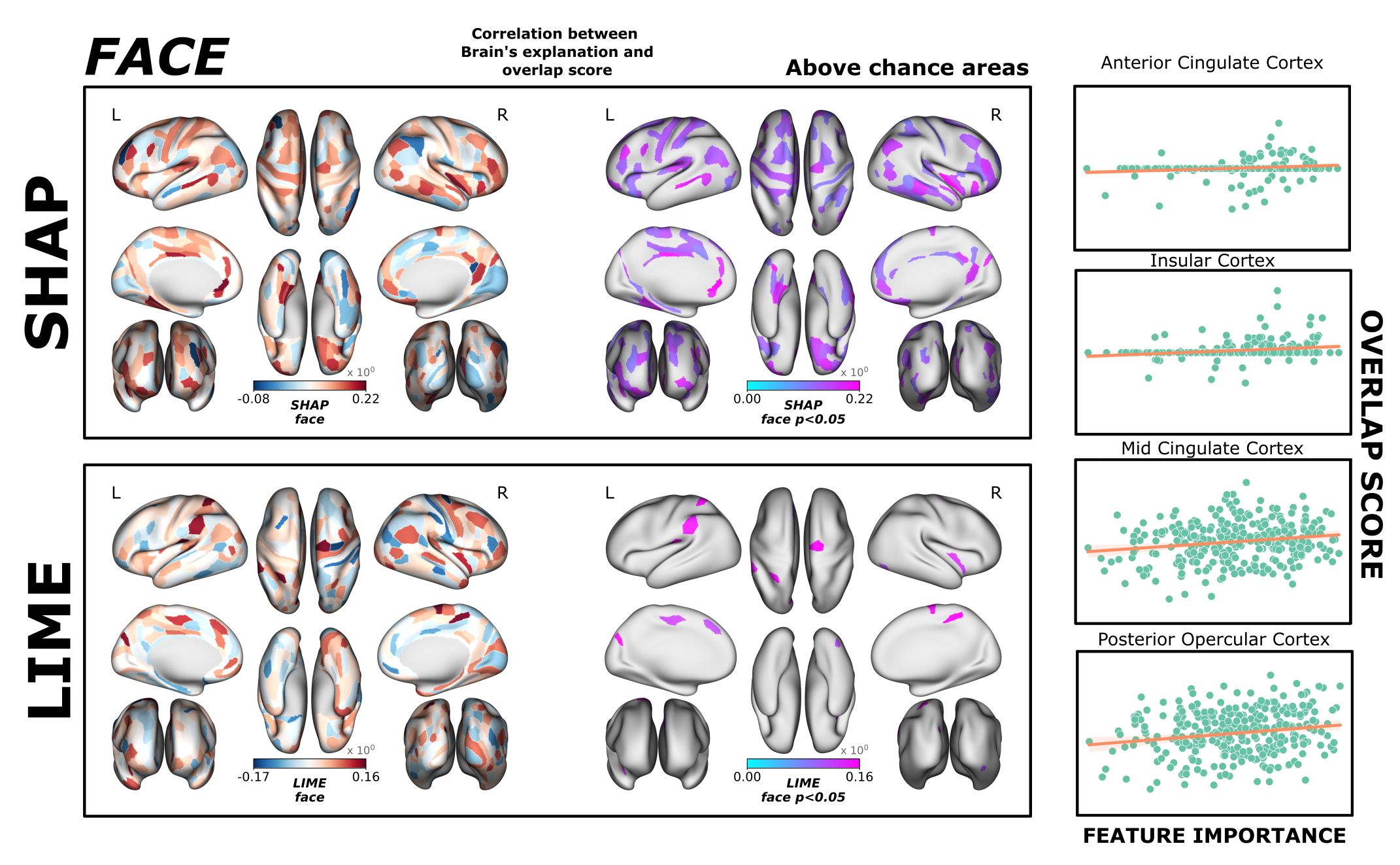}
  \caption{Area-wise correlation between Brain's explanation and overlap score.}
  \label{fig:eyetracking_corr_brains_fear}
\end{figure}

Since the advent of Deep Learning models for effective CV, comparing how humans and artificial intelligence systems process visual information has been an open research question. In this paper we tried to address such a complex issue, exploiting eye-tracking registration, fMRI data and XAI techniques. In particular, our challenge has been to understand if a specific brain pattern emerges whereas the attention of the human visual system and the CNN focus on the same spot within the presented visual stimuli, i.e. the frame of the Forrest Gump movie. Our hypothesis is that, if some region in the brain activates more when human eye-tracking and explainability cover the same pixels over the frame, it may represent a region of high similarity with CNNs with respect to the way they process visual information. This local (in the sense that we tried to look at a brain-region level) analysis produced a brain map for each of our 5 decoding model, highlighting the areas with a feature importance which correlates more with the degree of overlap between computer vision explanation over the frames and eye-tracking (Figure \ref{fig:eyetracking_corr_brains_fear}).

\begin{figure}[h!]
  \includegraphics[width=\linewidth]{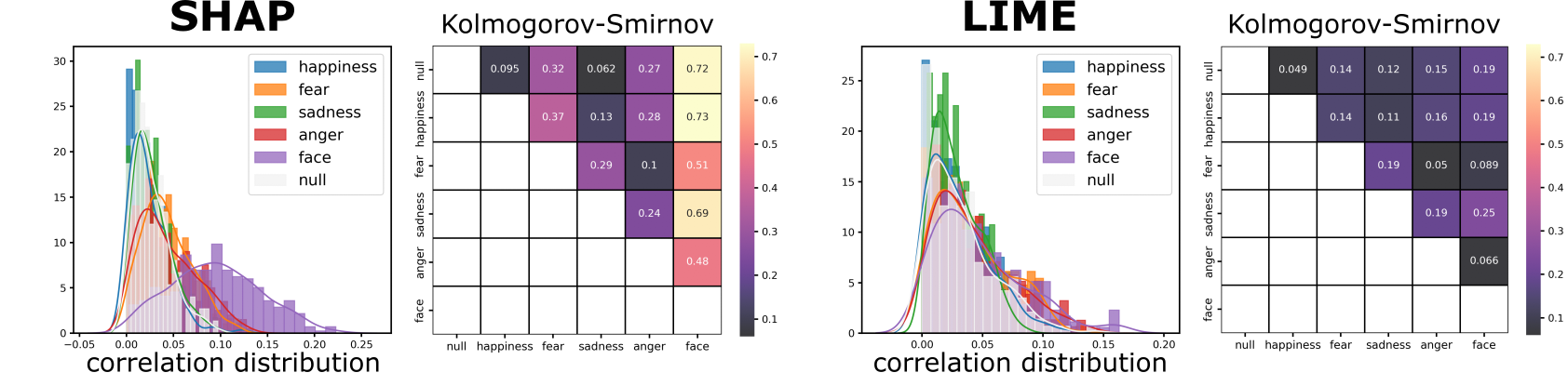}
  \caption{Correlation distributions per model and Kolmogorov-Smirnov distance. For what concern SHAP results, the face-related distribution shows globally higher correlation values, highlighting the locality of the visual information about faces, in contrast with the context-based nature of the emotional content of a visual stimulus.}
  \label{fig:eyetracking_correlation_distrib_permodel}
\end{figure}

The brain maps we found exhibit an intricate set of regions, different among models and not robust with respect to the choice of the explainer, making the results hard to discuss. However, we can move at a global whole-brain level in order to get some insights from this kind of analysis. Looking at the distribution of the correlation values for each brain region, displayed in Figures \ref{fig:eyetracking_correlation_distrib_permodel}, we can observe interesting phenomena, mostly regarding the results with SHAP. Despite the distributions of the emotion-related models seem to basically overlap, with no appreciable difference with respect to a null distribution, the values associated to the face model displayed higher correlation values. Such a result tells us that, globally, there are some brain regions with a non-negligible tendency to activate the most when the computer vision and the human being focus on the same part of the visual stimulus. At a first level, this finding can be interpreted as a natural consequence of the multisensory nature of the emotional content. 
The face recognition task is a visual-only task, thus it is more probable to find correlations between visual attention and fMRI-based explanations.
Then, another level of interpretation of our results regards the context-based nature of the emotional content of visual stimuli. The face recognition task is basically driven by local features of the visual input. The presence of a face within a frame is not related to the context where the face is located. For what concerns emotions, it's clear that this kind of information about a stimulus is conveyed by the whole, not only by a limited detail. Under these assumptions, it's easier to understand why the information about the attentional match among humans and machines does not convey information about similarities among CNN and the brain visual pathways.
\begin{table}[h!]
\centering
\captionsetup{skip=10pt} 
\caption{Most signigficative areas in terms of correlation among shap values and overlap between eyetracking and explainability over the frames.}
\begin{adjustbox}{width=\textwidth}
\begin{tabular}{|
>{\columncolor[HTML]{EFEFEF}}l |
>{\columncolor[HTML]{EFEFEF}}l 
>{\columncolor[HTML]{EFEFEF}}l |
>{\columncolor[HTML]{EFEFEF}}l 
>{\columncolor[HTML]{EFEFEF}}l |}
\hline
\cellcolor[HTML]{FFFFFF}\textbf{}                                     & \multicolumn{2}{c|}{\cellcolor[HTML]{FFFFFF}\textbf{SHAP}}                                                 & \multicolumn{2}{c|}{\cellcolor[HTML]{FFFFFF}\textbf{LIME}}                                                 \\ \hline
\cellcolor[HTML]{FFFFFF}\textbf{Model}                                & \cellcolor[HTML]{FFFFFF}\textbf{Macro Area}                 & \cellcolor[HTML]{FFFFFF}\textbf{Correlation} & \cellcolor[HTML]{FFFFFF}\textbf{Macro Area}                 & \cellcolor[HTML]{FFFFFF}\textbf{Correlation} \\
\cellcolor[HTML]{EFEFEF}                                              & Early Auditory Cortex                                       & 0.119                                        & Premotor Cortex                                             & 0.123                                        \\
\cellcolor[HTML]{EFEFEF}                                              &                                                             &                                              & Orbital and Polar Frontal Cortex                            & 0.110                                        \\
\cellcolor[HTML]{EFEFEF}                                              &                                                             &                                              & Medial Temporal Cortex                                      & 0.103                                        \\
\multirow{-4}{*}{\cellcolor[HTML]{EFEFEF}\textit{\textbf{Happiness}}} &                                                             &                                              & Mid Cingulate Cortex                                        & 0.098                                        \\ \hline
\cellcolor[HTML]{FFFFFF}                                              & \cellcolor[HTML]{FFFFFF}TPOJ & \cellcolor[HTML]{FFFFFF}0.119                & \cellcolor[HTML]{FFFFFF}Dorsal Stream Visual Cortex         & \cellcolor[HTML]{FFFFFF}0.134                \\
\cellcolor[HTML]{FFFFFF}                                              & \cellcolor[HTML]{FFFFFF}Insular Cortex                      & \cellcolor[HTML]{FFFFFF}0.119                & \cellcolor[HTML]{FFFFFF}Early Auditory Cortex               & \cellcolor[HTML]{FFFFFF}0.109                \\
\cellcolor[HTML]{FFFFFF}                                              & \cellcolor[HTML]{FFFFFF}Superior Parietal Cortex            & \cellcolor[HTML]{FFFFFF}0.116                & \cellcolor[HTML]{FFFFFF}Nucleus Accumbens                   & \cellcolor[HTML]{FFFFFF}0.108                \\
\cellcolor[HTML]{FFFFFF}                                              & \cellcolor[HTML]{FFFFFF}Neighboring Visual Areas            & \cellcolor[HTML]{FFFFFF}0.111                & \cellcolor[HTML]{FFFFFF}Premotor Cortex                     & \cellcolor[HTML]{FFFFFF}0.107                \\
\multirow{-5}{*}{\cellcolor[HTML]{FFFFFF}\textit{\textbf{Fear}}}      & \cellcolor[HTML]{FFFFFF}Ventral Stream Visual Cortex        & \cellcolor[HTML]{FFFFFF}0.107                & \cellcolor[HTML]{FFFFFF}Mid Cingulate Cortex                & \cellcolor[HTML]{FFFFFF}0.100                \\ \hline
\cellcolor[HTML]{EFEFEF}                                              & Orbital and Polar Frontal Cortex                            & 0.0966                                       & Auditory Association Cortex                                 & 0.086                                        \\
\cellcolor[HTML]{EFEFEF}                                              & Posterior Cingulate Cortex                                  & 0.0936                                       &                                                             &                                              \\
\multirow{-3}{*}{\cellcolor[HTML]{EFEFEF}\textit{\textbf{Sadness}}}   & Dorsolateral Prefrontal Cortex                              & 0.0836                                       &                                                             &                                              \\ \hline
\cellcolor[HTML]{FFFFFF}                                              & \cellcolor[HTML]{FFFFFF}Dorsal Stream Visual Cortex         & \cellcolor[HTML]{FFFFFF}0.134                & \cellcolor[HTML]{FFFFFF}Insular Cortex                      & \cellcolor[HTML]{FFFFFF}0.142                \\
\cellcolor[HTML]{FFFFFF}                                              & \cellcolor[HTML]{FFFFFF}Somatosensory and Motor Cortex      & \cellcolor[HTML]{FFFFFF}0.125                & \cellcolor[HTML]{FFFFFF}Anterior Cingulate Cortex           & \cellcolor[HTML]{FFFFFF}0.134                \\
\cellcolor[HTML]{FFFFFF}                                              & \cellcolor[HTML]{FFFFFF}Lateral Temporal Cortex             & \cellcolor[HTML]{FFFFFF}0.123                & \cellcolor[HTML]{FFFFFF}Premotor Cortex                     & \cellcolor[HTML]{FFFFFF}0.132                \\
\multirow{-4}{*}{\cellcolor[HTML]{FFFFFF}\textit{\textbf{Anger}}}     & \cellcolor[HTML]{FFFFFF}Posterior Cingulate Cortex          & \cellcolor[HTML]{FFFFFF}0.122                & \cellcolor[HTML]{FFFFFF}TPOJ & \cellcolor[HTML]{FFFFFF}0.124                \\ \hline
\cellcolor[HTML]{EFEFEF}                                              & Anterior Cingulate                                          & 0.223                                        & Posterior Opercular Cortex                                  & 0.163                                        \\
\cellcolor[HTML]{EFEFEF}                                              & Insular Cortex                                              & 0.218                                        & Mid Cingulate Cortex                                        & 0.161                                        \\
\cellcolor[HTML]{EFEFEF}                                              & Medial Temporal Cortex                                      & 0.217                                        & Superior Parietal Cortex                                    & 0.161                                        \\
\cellcolor[HTML]{EFEFEF}                                              & Posterior Cingulate Cortex                                  & 0.204                                        & Dorsal Stream Visual Cortex                                 & 0.155                                        \\
\multirow{-5}{*}{\cellcolor[HTML]{EFEFEF}\textit{\textbf{Face}}}      & Ventral Stream Visual Cortex                                & 0.199                                        &                                                             &                                              \\ \hline
\end{tabular}
\end{adjustbox}
\end{table}

\section{Conclusions}
To decode emotional states from visual stimuli in an ecological environment represents a challenging problem, in terms of understanding how the human brain processes such information and with respect to the attempt of building artificial intelligence systems able to replicate human perceptive and cognitive capabilities. In this paper we explored the possibility to retrieve information about emotional content of sensory input exploiting ML models. Moreover, we exploited XAI techniques in order to unveil how the brain processes emotional states. Through a computer vision-based emotion decoding analysis, we explored state-of-the-art techniques regarding the emotion classification within a set of complex visual stimuli, by showing that high-level semantic information can be successfully decoded by a CV model with high performance. 

For the brain decoding problem, we obtained high performance results in terms of classification. Moreover, the XAI analysis we performed outlined the existence of a fundamental network which processes the emotional content of a complex sensory stimulus, with various levels of processing stages, from a lower-level sensory elaboration to a higher-level cognitive processing. These findings confirm a constructionist vision regarding how the brain creates emotional feelings, without the involvement of emotion-specific regions. However, the evidence put forth in this research is not compelling enough to invalidate the locationist theory. In fact, our machine learning-based analysis has been designed to discriminate the presence or absence of a single emotion at times, not accounting for the discrimination among different emotions. The emotion-wise network that we observe tells us that an emotion-independent set of brain regions processes the whole spectrum of emotional content of a stimulus, but does not exclude that some specific areas can be involved in the elaboration of specific emotions.

Applying XAI techniques also to the computer vision models, we tried to compare the attentional mechanisms beyond the elaboration of emotional content of a stimulus in CNNs and human beings. This comparative analysis was aimed at eliciting specific patterns of activations in the brain whereas the human vision system and the artificial intelligence focus their attention in the same spot within the visual input. We found that it's possible to define some kind of coherent patterns of correlation with the attention match between humans and CNN only in the context of face recognition, because the evaluation of emotional context is intrinsically a context-based process.
In other words, this kind of analysis aims to exploit an attentional similarity between CNN and human vision, which can be found in face recognition, but not in emotion decoding tasks, in order to define a relation about brain regions which show analogy with CNNs.

Through our analysis we addressed the emotion decoding problem from an XAI perspective, exploring innovative tools to explore the way the human brain and CNNs process emotional sensory stimuli. We also tried to look at the bridge between CNNs and the human visual system from a different perspective, exploiting the attentional mechanisms incorporated within these two systems in order to identify which portions of the brain show some kind of synchronized behavior. Our work has been focused on single CNN architecture, namely the EfficientNet B0 architecture. Further works could explore the relations between more complex models and the human brain. However, the transfer learning procedure we adopted provides to the CNN general visual features, ensuring a strong reliability of our results.

A further analysis that would extend our observations is the representation similarity analysis (RSA) between the fMRI brain activity and the emotion decoding with CNN ~\cite{kriegeskorte_deep_2015}. RSA would be able to measure the similarity of processing between biological and artificial visual systems, as well as quantifying the differences of their inner representations.

\section*{Acknowledgement}
We extend our heartfelt gratitude to Cristina Solinas for her invaluable insights and interpretations concerning the neuroscientific aspects of our results.

%
%
%
\pagebreak
 \bibliographystyle{splncs04}
 \bibliography{bibliography}

\begin{thebibliography}{10}
\providecommand{\url}[1]{\texttt{#1}}
\providecommand{\urlprefix}{URL }
\providecommand{\doi}[1]{https://doi.org/#1}

\bibitem{ahmed_systematic_2023}
Ahmed, N., Aghbari, Z.A., Girija, S.: A systematic survey on multimodal emotion recognition using learning algorithms. Intelligent Systems with Applications  \textbf{17},  200171 (Feb 2023). \doi{10.1016/j.iswa.2022.200171}

\bibitem{akamatsu_perceived_2021}
Akamatsu, Y., Harakawa, R., Ogawa, T., Haseyama, M.: Perceived {Image} {Decoding} {From} {Brain} {Activity} {Using} {Shared} {Information} of {Multi}-{Subject} {fMRI} {Data}. IEEE Access  \textbf{9},  26593--26606 (2021). \doi{10.1109/ACCESS.2021.3057800}, \url{https://ieeexplore.ieee.org/document/9349437/}

\bibitem{alexander-bloch_testing_2018}
Alexander-Bloch, A.F., Shou, H., Liu, S., Satterthwaite, T.D., Glahn, D.C., Shinohara, R.T., Vandekar, S.N., Raznahan, A.: On testing for spatial correspondence between maps of human brain structure and function. NeuroImage  \textbf{178},  540--551 (Sep 2018). \doi{10.1016/j.neuroimage.2018.05.070}

\bibitem{arrieta_survey}
{Barredo Arrieta}, A., Díaz-Rodríguez, N., {Del Ser}, J., Bennetot, A., Tabik, S., Barbado, A., Garcia, S., Gil-Lopez, S., Molina, D., Benjamins, R., Chatila, R., Herrera, F.: Explainable artificial intelligence (xai): Concepts, taxonomies, opportunities and challenges toward responsible ai. Information Fusion  \textbf{58},  82--115 (2020)

\bibitem{barrett_analyzing_2019}
Barrett, D.G., Morcos, A.S., Macke, J.H.: Analyzing biological and artificial neural networks: challenges with opportunities for synergy? Current Opinion in Neurobiology  \textbf{55},  55--64 (Apr 2019). \doi{10.1016/j.conb.2019.01.007}

\bibitem{barrett_chapter_2009}
Barrett, L.F., Bliss‐Moreau, E.: Chapter 4 {Affect} as a {Psychological} {Primitive}. In: Advances in {Experimental} {Social} {Psychology}, vol.~41, pp. 167--218. Elsevier (2009). \doi{10.1016/S0065-2601(08)00404-8}

\bibitem{baucom_decoding_2012}
Baucom, L.B., Wedell, D.H., Wang, J., Blitzer, D.N., Shinkareva, S.V.: Decoding the neural representation of affective states. NeuroImage  \textbf{59}(1),  718--727 (Jan 2012). \doi{10.1016/j.neuroimage.2011.07.037}

\bibitem{bodria_benchmarking_2021}
Bodria, F., Giannotti, F., Guidotti, R., Naretto, F., Pedreschi, D., Rinzivillo, S.: Benchmarking and {Survey} of {Explanation} {Methods} for {Black} {Box} {Models} (Feb 2021). \doi{10.48550/arXiv.2102.13076}, arXiv:2102.13076 [cs]

\bibitem{bud_craig_how_2009}
(Bud)~Craig, A.D.: How do you feel — now? {The} anterior insula and human awareness. Nature Reviews Neuroscience  \textbf{10}(1),  59--70 (Jan 2009). \doi{10.1038/nrn2555}

\bibitem{connolly_representation_2012}
Connolly, A.C., Guntupalli, J.S., Gors, J., Hanke, M., Halchenko, Y.O., Wu, Y.C., Abdi, H., Haxby, J.V.: The {Representation} of {Biological} {Classes} in the {Human} {Brain}. The Journal of Neuroscience  \textbf{32}(8),  2608--2618 (Feb 2012). \doi{10.1523/JNEUROSCI.5547-11.2012}

\bibitem{conway_organization_2018}
Conway, B.R.: The {Organization} and {Operation} of {Inferior} {Temporal} {Cortex}. Annual Review of Vision Science  \textbf{4}(1),  381--402 (Sep 2018). \doi{10.1146/annurev-vision-091517-034202}

\bibitem{cox_software_1997}
Cox, R.W., Hyde, J.S.: Software tools for analysis and visualization of {fMRI} data. NMR in Biomedicine  \textbf{10}(4-5),  171--178 (Jun 1997). \doi{10.1002/(SICI)1099-1492(199706/08)10:4/5<171::AID-NBM453>3.0.CO;2-L}

\bibitem{du_spatio-temporal_2021}
Du, Z., Wu, S., Huang, D., Li, W., Wang, Y.: Spatio-{Temporal} {Encoder}-{Decoder} {Fully} {Convolutional} {Network} for {Video}-{Based} {Dimensional} {Emotion} {Recognition}. IEEE Transactions on Affective Computing  \textbf{12}(3),  565--578 (Jul 2021). \doi{10.1109/TAFFC.2019.2940224}

\bibitem{fan2011involvement}
Fan, J., Gu, X., Liu, X., Guise, K.G., Park, Y., Martin, L., de~Marchena, A., Tang, C.Y., Minzenberg, M.J., Hof, P.R.: Involvement of the anterior cingulate and frontoinsular cortices in rapid processing of salient facial emotional information. Neuroimage  \textbf{54}(3),  2539--2546 (2011)

\bibitem{farahani_explainable_2022}
Farahani, F.V., Fiok, K., Lahijanian, B., Karwowski, W., Douglas, P.K.: Explainable {AI}: {A} review of applications to neuroimaging data. Frontiers in Neuroscience  \textbf{16},  906290 (Dec 2022). \doi{10.3389/fnins.2022.906290}

\bibitem{fasel_automatic_2003}
Fasel, B., Luettin, J.: Automatic facial expression analysis: a survey. Pattern Recognition  (2003)

\bibitem{firat_deep_2014}
Firat, O., Oztekin, L., Vural, F.T.Y.: Deep learning for brain decoding. In: 2014 {IEEE} {International} {Conference} on {Image} {Processing} ({ICIP}). pp. 2784--2788. IEEE, Paris, France (Oct 2014). \doi{10.1109/ICIP.2014.7025563}

\bibitem{glasser_multi-modal_2016}
Glasser, M.F., Coalson, T.S., Robinson, E.C., Hacker, C.D., Harwell, J., Yacoub, E., Ugurbil, K., Andersson, J., Beckmann, C.F., Jenkinson, M., Smith, S.M., Van~Essen, D.C.: A multi-modal parcellation of human cerebral cortex. Nature  \textbf{536}(7615),  171--178 (Aug 2016). \doi{10.1038/nature18933}

\bibitem{glorot_understanding_nodate}
Glorot, X., Bengio, Y.: Understanding the difﬁculty of training deep feedforward neural networks

\bibitem{gunes_emotion_2011}
Gunes, H., Schuller, B., Pantic, M., Cowie, R.: Emotion representation, analysis and synthesis in continuous space: {A} survey. In: Face and {Gesture} 2011. pp. 827--834. IEEE, Santa Barbara, CA, USA (Mar 2011). \doi{10.1109/FG.2011.5771357}

\bibitem{haines_using_2019}
Haines, N., Southward, M.W., Cheavens, J.S., Beauchaine, T., Ahn, W.Y.: Using computer-vision and machine learning to automate facial coding of positive and negative affect intensity. PLOS ONE  \textbf{14}(2),  e0211735 (Feb 2019). \doi{10.1371/journal.pone.0211735}

\bibitem{hanke_simultaneous_2016}
Hanke, M., Adelhöfer, N., Kottke, D., Iacovella, V., Sengupta, A., Kaule, F.R., Nigbur, R., Waite, A.Q., Baumgartner, F.J., Stadler, J.: Simultaneous {fMRI} and eye gaze recordings during prolonged natural stimulation - a studyforrest extension (Mar 2016). \doi{10.1101/046581}

\bibitem{hanke_high-resolution_2014}
Hanke, M., Baumgartner, F.J., Ibe, P., Kaule, F.R., Pollmann, S., Speck, O., Zinke, W., Stadler, J.: A high-resolution 7-{Tesla} {fMRI} dataset from complex natural stimulation with an audio movie. Scientific Data  \textbf{1}(1),  140003 (May 2014). \doi{10.1038/sdata.2014.3}

\bibitem{haxby_multivariate_2012}
Haxby, J.V.: Multivariate pattern analysis of {fMRI}: {The} early beginnings. NeuroImage  \textbf{62}(2),  852--855 (Aug 2012). \doi{10.1016/j.neuroimage.2012.03.016}

\bibitem{heeger_what_2002}
Heeger, D.J., Ress, D.: What does {fMRI} tell us about neuronal activity? Nature Reviews Neuroscience  \textbf{3}(2),  142--151 (Feb 2002). \doi{10.1038/nrn730}

\bibitem{heinzle_multivariate_2012}
Heinzle, J., Anders, S., Bode, S., Bogler, C., Chen, Y., Cichy, R., Hackmack, K., Kahnt, T., Kalberlah, C., Reverberi, C., Soon, C., Tusche, A., Weygandt, M., Haynes, J.D.: Multivariate decoding of {fMRI} data: {Towards} a content-based cognitive neuroscience. e-Neuroforum  \textbf{18}(1),  1--16 (Mar 2012). \doi{10.1007/s13295-012-0026-9}

\bibitem{hornak_changes_2003}
Hornak, J.: Changes in emotion after circumscribed surgical lesions of the orbitofrontal and cingulate cortices. Brain  \textbf{126}(7),  1691--1712 (Apr 2003). \doi{10.1093/brain/awg168}

\bibitem{jabbi_common_2008}
Jabbi, M., Bastiaansen, J., Keysers, C.: A {Common} {Anterior} {Insula} {Representation} of {Disgust} {Observation}, {Experience} and {Imagination} {Shows} {Divergent} {Functional} {Connectivity} {Pathways}. PLoS ONE  \textbf{3}(8),  e2939 (Aug 2008). \doi{10.1371/journal.pone.0002939}

\bibitem{jenkinson_fsl_2012}
Jenkinson, M., Beckmann, C.F., Behrens, T.E., Woolrich, M.W., Smith, S.M.: {FSL}. NeuroImage  \textbf{62}(2),  782--790 (Aug 2012). \doi{10.1016/j.neuroimage.2011.09.015}

\bibitem{kingma_adam_2017}
Kingma, D.P., Ba, J.: Adam: {A} {Method} for {Stochastic} {Optimization} (Jan 2017), arXiv:1412.6980 [cs]

\bibitem{ko_brief_2018}
Ko, B.: A {Brief} {Review} of {Facial} {Emotion} {Recognition} {Based} on {Visual} {Information}. Sensors  \textbf{18}(2), ~401 (Jan 2018). \doi{10.3390/s18020401}

\bibitem{k_deep_2015}
Koyamada, S., Shikauchi, Y., Nakae, K., Koyama, M., Ishii, S.: Deep learning of {fMRI} big data: a novel approach to subject-transfer decoding (Jan 2015), arXiv:1502.00093 [cs, q-bio, stat]

\bibitem{kragel_multivariate_2015}
Kragel, P.A., LaBar, K.S.: Multivariate neural biomarkers of emotional states are categorically distinct. Social Cognitive and Affective Neuroscience  \textbf{10}(11),  1437--1448 (Nov 2015). \doi{10.1093/scan/nsv032}

\bibitem{kragel_decoding_2016}
Kragel, P.A., LaBar, K.S.: Decoding the {Nature} of {Emotion} in the {Brain}. Trends in cognitive sciences  \textbf{20}(6),  444--455 (Jun 2016). \doi{10.1016/j.tics.2016.03.011}

\bibitem{kriegeskorte_deep_2015}
Kriegeskorte, N.: Deep {Neural} {Networks}: {A} {New} {Framework} for {Modeling} {Biological} {Vision} and {Brain} {Information} {Processing}. Annual Review of Vision Science  \textbf{1}(1),  417--446 (Nov 2015). \doi{10.1146/annurev-vision-082114-035447}

\bibitem{kringelbach_functional_2004}
Kringelbach, M.: The functional neuroanatomy of the human orbitofrontal cortex: evidence from neuroimaging and neuropsychology. Progress in Neurobiology  \textbf{72}(5),  341--372 (Apr 2004). \doi{10.1016/j.pneurobio.2004.03.006}

\bibitem{alexnet}
Krizhevsky, A., Sutskever, I., Hinton, G.E.: Imagenet classification with deep convolutional neural networks. In: Pereira, F., Burges, C., Bottou, L., Weinberger, K. (eds.) Advances in Neural Information Processing Systems. vol.~25. Curran Associates, Inc. (2012)

\bibitem{kubilius_brain-decoding_2015}
Kubilius, J., Baeck, A., Wagemans, J., Op~De~Beeck, H.P.: Brain-decoding {fMRI} reveals how wholes relate to the sum of parts. Cortex  \textbf{72},  5--14 (Nov 2015). \doi{10.1016/j.cortex.2015.01.020}

\bibitem{labs_portrayed_2015}
Labs, A., Reich, T., Schulenburg, H., Boennen, M., Mareike, G., Golz, M., Hartigs, B., Hoffmann, N., Keil, S., Perlow, M., Peukmann, A.K., Rabe, L.N., von Sobbe, F.R., Hanke, M.: Portrayed emotions in the movie "{Forrest} {Gump}". F1000Research  \textbf{4}, ~92 (Apr 2015). \doi{10.12688/f1000research.6230.1}

\bibitem{lee_fast_2022}
Lee, S., Bradlow, E.T., Kable, J.W.: Fast construction of interpretable whole-brain decoders. Cell Reports Methods  \textbf{2}(6),  100227 (Jun 2022). \doi{10.1016/j.crmeth.2022.100227}

\bibitem{lettieri_emotionotopy_2019}
Lettieri, G., Handjaras, G., Ricciardi, E., Leo, A., Papale, P., Betta, M., Pietrini, P., Cecchetti, L.: Emotionotopy in the human right temporo-parietal cortex. Nature Communications  \textbf{10}(1), ~5568 (Dec 2019). \doi{10.1038/s41467-019-13599-z}

\bibitem{liang_cross-subject_2020}
Liang, Y., Liu, B.: Cross-{Subject} {Commonality} of {Emotion} {Representations} in {Dorsal} {Motion}-{Sensitive} {Areas}. Frontiers in Neuroscience  \textbf{14},  567797 (Oct 2020). \doi{10.3389/fnins.2020.567797}, \url{https://www.frontiersin.org/article/10.3389/fnins.2020.567797/full}

\bibitem{lin_advancing_2023}
Lin, C., Bulls, L.S., Tepfer, L.J., Vyas, A.D., Thornton, M.A.: Advancing {Naturalistic} {Affective} {Science} with {Deep} {Learning}. Affective Science  \textbf{4}(3),  550--562 (Sep 2023). \doi{10.1007/s42761-023-00215-z}

\bibitem{lindquist_brain_2012}
Lindquist, K.A., Wager, T.D., Kober, H., Bliss-Moreau, E., Barrett, L.F.: The brain basis of emotion: {A} meta-analytic review. Behavioral and Brain Sciences  \textbf{35}(3),  121--143 (Jun 2012). \doi{10.1017/S0140525X11000446}

\bibitem{lindsay_convolutional_2021}
Lindsay, G.W.: Convolutional {Neural} {Networks} as a {Model} of the {Visual} {System}: {Past}, {Present}, and {Future}. Journal of Cognitive Neuroscience  \textbf{33}(10),  2017--2031 (Sep 2021)

\bibitem{lopes_facial_2017}
Lopes, A.T., De~Aguiar, E., De~Souza, A.F., Oliveira-Santos, T.: Facial expression recognition with {Convolutional} {Neural} {Networks}: {Coping} with few data and the training sample order. Pattern Recognition  \textbf{61},  610--628 (Jan 2017). \doi{10.1016/j.patcog.2016.07.026}

\bibitem{lundberg_unified_2017}
Lundberg, S.M., Lee, S.I.: A {Unified} {Approach} to {Interpreting} {Model} {Predictions}. In: Advances in {Neural} {Information} {Processing} {Systems}. vol.~30. Curran Associates, Inc. (2017)

\bibitem{Mellouk_Facial_2020}
Mellouk, W., Handouzi, W.: Facial emotion recognition using deep learning: review and insights. Procedia Computer Science  \textbf{175},  689--694 (2020). \doi{10.1016/j.procs.2020.07.101}

\bibitem{Miller_Explanation_2019}
Miller, T.: Explanation in artificial intelligence: {Insights} from the social sciences. Artificial Intelligence  \textbf{267},  1--38 (Feb 2019). \doi{10.1016/j.artint.2018.07.007}

\bibitem{mitchell2012conscious}
Mitchell, D.G., Greening, S.G.: Conscious perception of emotional stimuli: brain mechanisms. The Neuroscientist  \textbf{18}(4),  386--398 (2012)

\bibitem{morita2008role}
Morita, T., Itakura, S., Saito, D.N., Nakashita, S., Harada, T., Kochiyama, T., Sadato, N.: The role of the right prefrontal cortex in self-evaluation of the face: a functional magnetic resonance imaging study. Journal of cognitive neuroscience  \textbf{20}(2),  342--355 (2008)

\bibitem{morita_anterior_2014}
Morita, T., Tanabe, H.C., Sasaki, A.T., Shimada, K., Kakigi, R., Sadato, N.: The anterior insular and anterior cingulate cortices in emotional processing for self-face recognition. Social Cognitive and Affective Neuroscience  \textbf{9}(5),  570--579 (May 2014). \doi{10.1093/scan/nst011}

\bibitem{murphy_functional_2003}
Murphy, F.C., Nimmo-Smith, I., Lawrence, A.D.: Functional neuroanatomy of emotions: {A} meta-analysis. Cognitive, Affective, \& Behavioral Neuroscience  \textbf{3}(3),  207--233 (Sep 2003). \doi{10.3758/CABN.3.3.207}

\bibitem{pat_explainable_2023}
Pat, N., Wang, Y., Bartonicek, A., Candia, J., Stringaris, A.: Explainable machine learning approach to predict and explain the relationship between task-based {fMRI} and individual differences in cognition. Cerebral Cortex  \textbf{33}(6),  2682--2703 (Mar 2023). \doi{10.1093/cercor/bhac235}

\bibitem{Pikoulis_Leveraging_2021}
Pikoulis, I., Filntisis, P.P., Maragos, P.: Leveraging {Semantic} {Scene} {Characteristics} and {Multi}-{Stream} {Convolutional} {Architectures} in a {Contextual} {Approach} for {Video}-{Based} {Visual} {Emotion} {Recognition} in the {Wild}. In: 2021 16th {IEEE} {International} {Conference} on {Automatic} {Face} and {Gesture} {Recognition} ({FG} 2021). pp. 01--08. IEEE, Jodhpur, India (Dec 2021). \doi{10.1109/FG52635.2021.9666957}

\bibitem{ribeiro_why_2016}
Ribeiro, M.T., Singh, S., Guestrin, C.: "{Why} {Should} {I} {Trust} {You}?": {Explaining} the {Predictions} of {Any} {Classifier}. In: Proceedings of the 22nd {ACM} {SIGKDD} {International} {Conference} on {Knowledge} {Discovery} and {Data} {Mining}. pp. 1135--1144. {KDD} '16, Association for Computing Machinery, New York, NY, USA (Aug 2016). \doi{10.1145/2939672.2939778}

\bibitem{rolls2019cingulate}
Rolls, E.T.: The cingulate cortex and limbic systems for emotion, action, and memory. Brain Structure and Function  \textbf{224}(9),  3001--3018 (2019)

\bibitem{saxe_if_2021}
Saxe, A., Nelli, S., Summerfield, C.: If deep learning is the answer, what is the question? Nature Reviews Neuroscience  \textbf{22}(1),  55--67 (Jan 2021). \doi{10.1038/s41583-020-00395-8}

\bibitem{schrimpf_brain-score_2018}
Schrimpf, M., Kubilius, J., Hong, H., Majaj, N.J., Rajalingham, R., Issa, E.B., Kar, K., Bashivan, P., Prescott-Roy, J., Geiger, F., Schmidt, K., Yamins, D.L.K., DiCarlo, J.J.: Brain-{Score}: {Which} {Artificial} {Neural} {Network} for {Object} {Recognition} is most {Brain}-{Like}? preprint, Neuroscience (Sep 2018). \doi{10.1101/407007}

\bibitem{sengupta_studyforrest_2016}
Sengupta, A., Kaule, F.R., Guntupalli, J.S., Hoffmann, M.B., Häusler, C., Stadler, J., Hanke, M.: A studyforrest extension, retinotopic mapping and localization of higher visual areas. Scientific Data  \textbf{3}(1),  160093 (Oct 2016). \doi{10.1038/sdata.2016.93}

\bibitem{serengil_lightface_2020}
Serengil, S.I., Ozpinar, A.: {LightFace}: {A} {Hybrid} {Deep} {Face} {Recognition} {Framework}. In: 2020 {Innovations} in {Intelligent} {Systems} and {Applications} {Conference} ({ASYU}). pp.~1--5 (Oct 2020). \doi{10.1109/ASYU50717.2020.9259802}

\bibitem{shorten_survey_2019}
Shorten, C., Khoshgoftaar, T.M.: A survey on {Image} {Data} {Augmentation} for {Deep} {Learning}. Journal of Big Data  \textbf{6}(1), ~60 (Dec 2019). \doi{10.1186/s40537-019-0197-0}

\bibitem{stevens_anterior_2011}
Stevens, F.L.: Anterior {Cingulate} {Cortex}: {Unique} {Role} in {Cognition} and {Emotion}. J Neuropsychiatry Clin Neurosci  (2011)

\bibitem{van2011cortico}
Van~den Stock, J., Tamietto, M., Sorger, B., Pichon, S., Gr{\'e}zes, J., de~Gelder, B.: Cortico-subcortical visual, somatosensory, and motor activations for perceiving dynamic whole-body emotional expressions with and without striate cortex (v1). Proceedings of the National Academy of Sciences  \textbf{108}(39),  16188--16193 (2011)

\bibitem{tan_efficientnet_2019}
Tan, M., Le, Q.V.: {EfficientNet}: {Rethinking} {Model} {Scaling} for {Convolutional} {Neural} {Networks} (May 2019)

\bibitem{Thuseethan_EmoSeC_2022}
Thuseethan, S., Rajasegarar, S., Yearwood, J.: {EmoSeC}: {Emotion} recognition from scene context. Neurocomputing  \textbf{492},  174--187 (Jul 2022). \doi{10.1016/j.neucom.2022.04.019}

\bibitem{vytal_neuroimaging_2010}
Vytal, K., Hamann, S.: Neuroimaging {Support} for {Discrete} {Neural} {Correlates} of {Basic} {Emotions}: {A} {Voxel}-based {Meta}-analysis. Journal of Cognitive Neuroscience  \textbf{22}(12),  2864--2885 (Dec 2010). \doi{10.1162/jocn.2009.21366}

\bibitem{weaverdyck_tools_2020}
Weaverdyck, M.E., Lieberman, M.D., Parkinson, C.: Tools of the {Trade} {Multivoxel} pattern analysis in {fMRI}: a practical introduction for social and affective neuroscientists. Social Cognitive and Affective Neuroscience  \textbf{15}(4),  487--509 (Jun 2020). \doi{10.1093/scan/nsaa057}

\bibitem{yamins_using_2016}
Yamins, D.L.K., DiCarlo, J.J.: Using goal-driven deep learning models to understand sensory cortex. Nature Neuroscience  \textbf{19}(3),  356--365 (Mar 2016). \doi{10.1038/nn.4244}

\bibitem{yao_early_2007}
Yao, Y., Rosasco, L., Caponnetto, A.: On {Early} {Stopping} in {Gradient} {Descent} {Learning}. Constructive Approximation  \textbf{26}(2),  289--315 (Aug 2007). \doi{10.1007/s00365-006-0663-2}

\bibitem{yousefnezhad_supervised_2021}
Yousefnezhad, M., Selvitella, A., Han, L., Zhang, D.: Supervised {Hyperalignment} for multi-subject {fMRI} data alignment. IEEE Transactions on Cognitive and Developmental Systems  \textbf{13}(3),  475--490 (Sep 2021). \doi{10.1109/TCDS.2020.2965981}, \url{http://arxiv.org/abs/2001.02894}, arXiv:2001.02894 [cs, q-bio, stat]

\bibitem{zhuang_comprehensive_2021}
Zhuang, F., Qi, Z., Duan, K., Xi, D., Zhu, Y., Zhu, H., Xiong, H., He, Q.: A {Comprehensive} {Survey} on {Transfer} {Learning}. Proceedings of the IEEE  \textbf{109}(1),  43--76 (Jan 2021). \doi{10.1109/JPROC.2020.3004555}

\end{thebibliography}

\end{document}